\documentclass{article}
\usepackage[eatrack]{log_2023}	

\usepackage{booktabs}            
\usepackage{multirow}            
\usepackage{amsfonts}            
\usepackage{graphicx}            
\usepackage{duckuments}          
\usepackage{tabularx}
\usepackage{tikz-cd}   
\usepackage{algorithm} 
\usepackage[noend]{algpseudocode} 
\usepackage{enumitem}
\usepackage{graphicx} 
\usepackage{subfigure}
\usepackage{amsthm}
\usepackage{wrapfig}
\usepackage{adjustbox}
\usepackage{diagbox}

\DeclareMathOperator*{\argmin}{arg\,min}

\usepackage[numbers,compress,sort]{natbib}

\title{Geometric Instability of Graph Neural Networks on Large Graphs}

\author[E. Morris, H. Shen, W. Du, M. Sajjad, B. Shi]{%
Emily Morris\thanks{All authors have equal contribution.}\\
University of Cambridge\\
\email{efm40@cam.ac.uk}
\And
Haotian Shen\footnotemark[1]\\
University of Cambridge\\
\email{hs831@cam.ac.uk}
\And
Weiling Du\footnotemark[1]\\
University of Cambridge\\
\email{wd301@cam.ac.uk}
\And
Muhammad Hamza Sajjad\footnotemark[1]\\
University of Cambridge\\
\email{mhs57@cam.ac.uk}
\And
Borun Shi\footnotemark[1]\kern0.4em \thanks{Corresponding author.}\\
Neo4j\\
\email{brian.shi@neo4j.com}
}

\begin{document}

\maketitle

\begin{abstract}
We analyse the geometric instability of embeddings produced by graph neural networks (GNNs). Existing methods are only applicable for small graphs and lack context in the graph domain. We propose a simple, efficient and graph-native Graph Gram Index (GGI) to measure such instability which is invariant to permutation, orthogonal transformation, translation and order of evaluation. This allows us to study the varying instability behaviour of GNN embeddings on large graphs for both node classification and link prediction.
\end{abstract}

\newtheorem{definition}{Definition} 
\newtheorem{lemma}{Lemma} 
\newtheorem*{lemma*}{Lemma} 
\newtheorem*{proof*}{Proof}

\section{Introduction}

Graph representation learning~\cite{Hamilton:ud} has seem much recent success in solving problems over relational data. 
The importance of stochastic effects is evident in graph learning. Earlier methods such as DeepWalk~\cite{DeepWalk} and node2vec~\cite{N2V} leverage random walk. Graph Neural Networks (GNN)\cite{GNN} contain stochastic components such as sampling and batching which allow training models on large real-world graphs. From a theoretical point of view, adding randomness is proven to improve expressiveness and performance \cite{sato2021random}\cite{abboud:2021}. The impact of randomness common in standard machine learning frameworks also carries over to the graph domain, such as in gradient descent and dataset splits\cite{shchur2019pitfalls}. It is therefore important to understand how all the stochastic components together affect graph learning models at a granular level.

Stability of embeddings and models have been extensively studied outside of the domain of graphs\cite{Borah2021AreWE}\cite{gursoy2023alignment}\cite{antoniak-mimno-2018-evaluating}\cite{Piccard:2021}. Yet the practical impact of randomness on GNNs has been very rarely studied. ~\cite{klabunde2022prediction} recently empirically showed that final predictions given by GCN~\cite{Kipf:2017tc} and GAT\cite{Velickovic:2018we} vary significantly on individual nodes despite relatively stable overall accuracies. At a finer level, ~\cite{Schumacher:2020} showed the instability in embeddings produced by several shallow methods and GraphSAGE~\cite{Hamilton:2017tp} and ~\cite{Hacker:2022} specifically for node2vec. To the best of our knowledge, all previous studies are carried out on small graphs for node classification and stability indices from other domains are directly borrowed.

Additional motivations of understanding embedding geometric stability include improving reproducibility and reliability~\cite{Piccard:2021}, reducing retraining effort when embeddings are used by multiple downstream systems by measuring drift~\cite{TwHIN}, and understanding how a model works by studying its embedding space.

The main contributions of this paper are:

\begin{enumerate}[label=Section \arabic*, start=2, leftmargin=4.5em, topsep=0pt, itemsep=0pt]
  	\item We formalise the notion of geometric stability index for embeddings and examine existing methods.
 	\item We propose a simple, time and space efficient and graph-native stability index, the Graph Gram index (GGI). We motivate and show that GGI is invariant to permutation, orthogonal transformation, translation and the order of evaluation.
  	\item We show the geometric stability behaviour of several popular GNNs on graphs of varying sizes and levels of homophily for node classification and link prediction.
\end{enumerate}

\section{Geometric stability}

\textbf{Notation.}
Let $G = (V, E)$ denote a graph, $|V|$ the number of nodes, $|E|$ the number of relationships. Each node has an input feature in $\mathbb{R}^{n}$, and a final embedding in $\mathbb{R}^{d}$. For a given node at position $i$ in some ordered list, we refer to that node as $v_i$. We refer to the final embedding of that node as $z_i \in \mathbb{R}^{1\times d}$. Whenever we need to refer to the embedding of some node $v_i$ by name, we will abuse the notation slightly as $z_{v_i}$. Let $Z \in \mathbb{R}^{|V| \times d}$ denote the embeddings of all nodes in the graph. Since we are concerned with multiple embeddings produced from different configurations, let superscript $Z^k$ denote the embeddings produced from configuration $k$, unless otherwise stated. Let $N$ denote the overall number of configurations.

Stability indices that have been used to evaluate the geometric stability of embeddings of graph models are either borrowed from the NLP domain~\cite{Schumacher:2020}, or from analytic topology ~\cite{Hacker:2022}. 
We give the definitions of one similarity index from each area and refer to the remaining ones in Appendix ~\ref{app:indices}.

\textbf{Second-order cosine.} Given two embedding matrices $Z^l$, $Z^m$. For each node $v_i$, defined an ordered list $\{u_1, \ldots., u_K\} = \mathcal{N}_k^l(v_i) \cup \mathcal{N}_k^m(v_i)$ where $\mathcal{N}_k^l(v_i)$ is the k nearest neighbour of node $i$ under configuration $l$. Let $s^l(v_i)$ denote a vector about $v_i$ where the $j^{th}$ entry is $s_j^l(v_i) = cossim(z_i^l, z_{u_j}^l)$. Similarly we can define $s^m(v_i)$. Let $s_k^{cos[l,m]}(z_i^l, z_i^m) = cossim(s_{v_i}^l, s_{v_i}^m)$. Cosine distance are sometimes equivalently used. The second order cosine similarity of $N$ configurations is then averaging over all nodes and all configuration pairs, i.e $\frac{1}{N^2 \times |V|} \sum_{l, m \in N \times N} \sum_{i \in V} s_k^{cos[l,m]}(z_i^l, z_i^m)$. This stability index was originally used to detect words semantic shifts over time~\cite{hamilton2016cultural}.

\textbf{Wasserstein distance.} Given two embedding matrices $Z^l$, $Z^m$, the Wasserstein distance between them is $W(Z^l, Z^m) = (\inf_{\eta : V \rightarrow V} \sum_{i \in V} \lVert z_i - z_{\eta(i)} \rVert^2_2)^{1/2}$. Where $\inf_{\eta: V \rightarrow V}$ is over all bijections between nodes. The average across all possible configuration pairs are taken as the stability index.

While these indices have shown preliminary instability results, there are several improvements to be made. First all of, most of the existing similarity indices are inefficient to compute, evident with the common \textit{averaging over all 
nodes and all configuration pairs}. Finding k nearest neighbours (kNN) itself is a costly operation\cite{johnson2019billion}.  
Opting for approximate kNN\cite{malkov2018efficient} introduces confounding randomness. Several of the neighbour-based indices contain hyperparameters of their own (\textit{k}), which for the same reason should ideally be avoided.
We give the full time and space complexity in Appendix~\ref{app:complexities}.

For any index that directly compares embeddings across configurations (i.e to compare $z_i^l$, $z_i^m$), an alignment step such as the solving the orthogonal Procrustes problem\cite{procrustes} (definition in Appendix~\ref{app:procrustes}) needs to be done first.
While embedded node neighbours can still carry important information, the interpretation of semantically similar words no longer applies. Indices from analytic topology (such as Wasserstein distance) are originally used to measure the overlap of probability distributions over metric space, which again does not map to any natural interpretation for graph embedding. Same drawback applies for Hausdorff distance (we elaborate more in Appendix~\ref{app:hausdorff} ).

We propose that a stability index for embeddings produced by graph machine learning models should ideally be time and space efficient to compute, free of hyperparameters of its own, and have an intuitive and simple interpretation in the context of graphs. Having such an index enables easier analysis of embeddings produced by graph models on large real-world graphs.

\section{Graph Gram Index}

We propose a simple index called the Graph Gram Index (GGI). 

\begin{algorithm}
\caption*{Graph Gram Index}\label{alg:GGI}
\begin{algorithmic}
\Procedure{GGI:}{$Z^1 \times (Z^2 \times (\dots Z^{N-1} \times Z^N$}
    \State \textbf{Input:} $Z^1, \dots, Z^N$
    \State \textbf{Output:} Stability index $s \in [0, 1]$
    \For{$l \in [1, \dots, N]$}
    	\State Center and normalize $Z^l$
	\vspace{0.8ex}
        \State $S^l = A \circ Z^l \space {Z^l}^{T}$
        \State $s^l = \frac{1}{2|E|}\sum_{i,j \in |V| \times |V|}S^l[i,j]$
    \EndFor
    \State $s = \text{std}(s^l: l \in [1, \dots, N)$
    \State \text{where A is the adjacency matrix and $\circ$ is the hadamard product.}
\EndProcedure
\end{algorithmic}
\end{algorithm}

\captionsetup[figure]{skip=0pt}  

\begin{figure}
\centering
\begin{tikzcd}
& Z^1 \dots Z^N \arrow{d}{\text{Summary structure}}  \arrow{rr}{\forall l, m \in N \times N} &
	& (Z^l, Z^m) \arrow{rr}{\text{Align+Compare}}  &
	& s^{l,m} \arrow{dd}{\text{Average}} \\
& S^1 \dots S^N \arrow{d}{\text{Pool}} \\
& s^1 \dots s^N \arrow{rrrr}{\text{std}} & & & & s \\
\end{tikzcd}
\caption{Commutative diagram showing the approach of GGI. All previous methods follow the bottom-left path whereas we use the top-right, which scales better to large graphs and offers a flexible framework to create other indices.}
\label{fig:commutativediagram}
\end{figure}
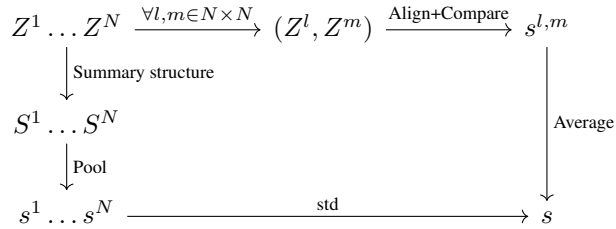

$Z^l{Z^l}^T$ is indeed the Gram matrix, which contains all covariance information between $z_i^l$, $z_j^l$. Using the Gram matrix avoids solving the Procrustes alignment step which would be required if we calculate $z_i^lz_j^m$. Intuitively, this is achieved because we \textit{compare within one set of embeddings to generate a structure summary, and compare the summaries across configurations,} illustrated in Figure~\ref{fig:commutativediagram}

An analogous approach underpins Centered Kernel Alignment\cite{CKA}(Appendix~\ref{app:cka}), a \textit{similarity} measure to compare internal representations across general neural networks (of any domain).

Applying a hadamard product of the adjacency matrix over the Gram matrix means that we only capture the node pairs that are actual edges in the graph. Intuitively, a stable model produces embeddings whose node pairs that correspond to an edge should remain similar, in their respective embedding spaces, \textit{up to any equivariant transformation satisfied by the specific model}. Note that a node pair corresponding to an edge could potentially lie far apart in the embedding space, which previous stability indices would not consider.
The framework of GGI is flexible enough to allow various extension, we list some future work in Appendix~\ref{app:futurework}.

\subsection{Invariance properties}

To the best of our knowledge, there is no prior study about the geometric invariances\cite{maron2019universality}\cite{bronstein2021geometric} a stability index should satisfy. It is important to define such invariances, because the meaningful geometric instability any index captures should be the ones that consumers of the embeddings (final GNN layer or downstream model\cite{lecun91}\cite{chen93}) are not equivariant to. At the same time, it should ideally not be only invariant to an overly specific set of transformations that few models satisfy. 

Hence, we propose that it is desirable to be invariant to node permutation, orthogonal transformation and translation. In addition it should be invariant to the order of evaluating a given set of configurations. kNN-Jaccard and second-order cossim for example are not invariant to permutation due to it's own nondeterminism from sampling. We list some future work in Appendix~\ref{app:futurework}.

 \begin{lemma*}
GGI is invariant to node permutation, orthogonal transformation and translation of embeddings, and order of evaluation.
\label{lemma:cgeainvariance}
\end{lemma*}

Proof follows from simple applications of properties of matrix multiplication and isometries which we include in Appendix~\ref{app:proof}.

\section{Experiments}
\textbf{Node classification.}
We evaluate four popular baseline GNN models, GCN, GraphSAGE, GAT, GIN\cite{GIN} on cora and ogbn-arxiv\cite{ogb}. Additionally, we evaluate the directional version of the models\cite{rossi2023edge} on roman-empire (a heterophilic dataset)\cite{platonov2023critical}. We measure the stability of each model across 30 different random seeds.
Full training setup is detailed in Appendix~\ref{app:experimentsetup}.

\begin{figure}[htbp]
    \centering
    \subfigure[Cora]{\includegraphics[width=0.48\linewidth]{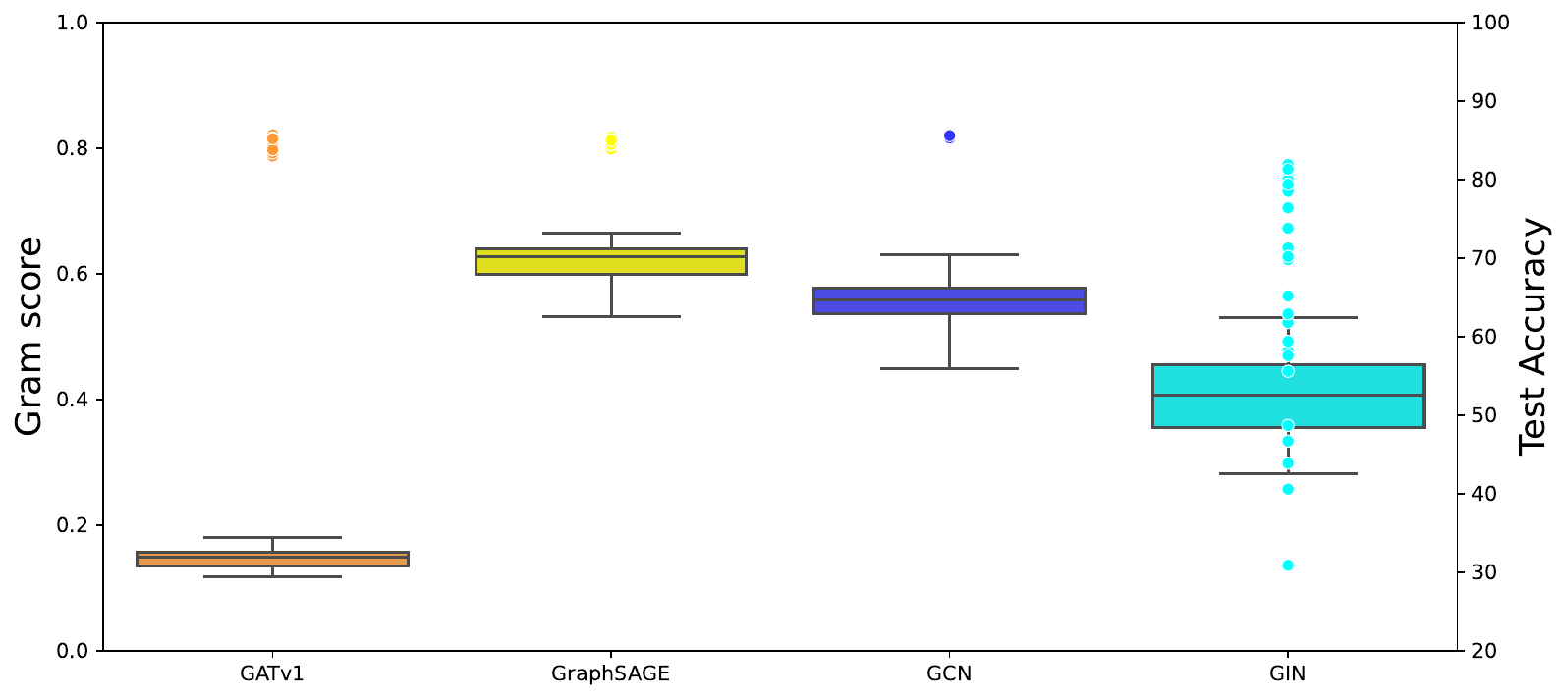}\label{fig:subfig3}}
    \subfigure[ogbn-arxiv]{\includegraphics[width=0.48\linewidth]{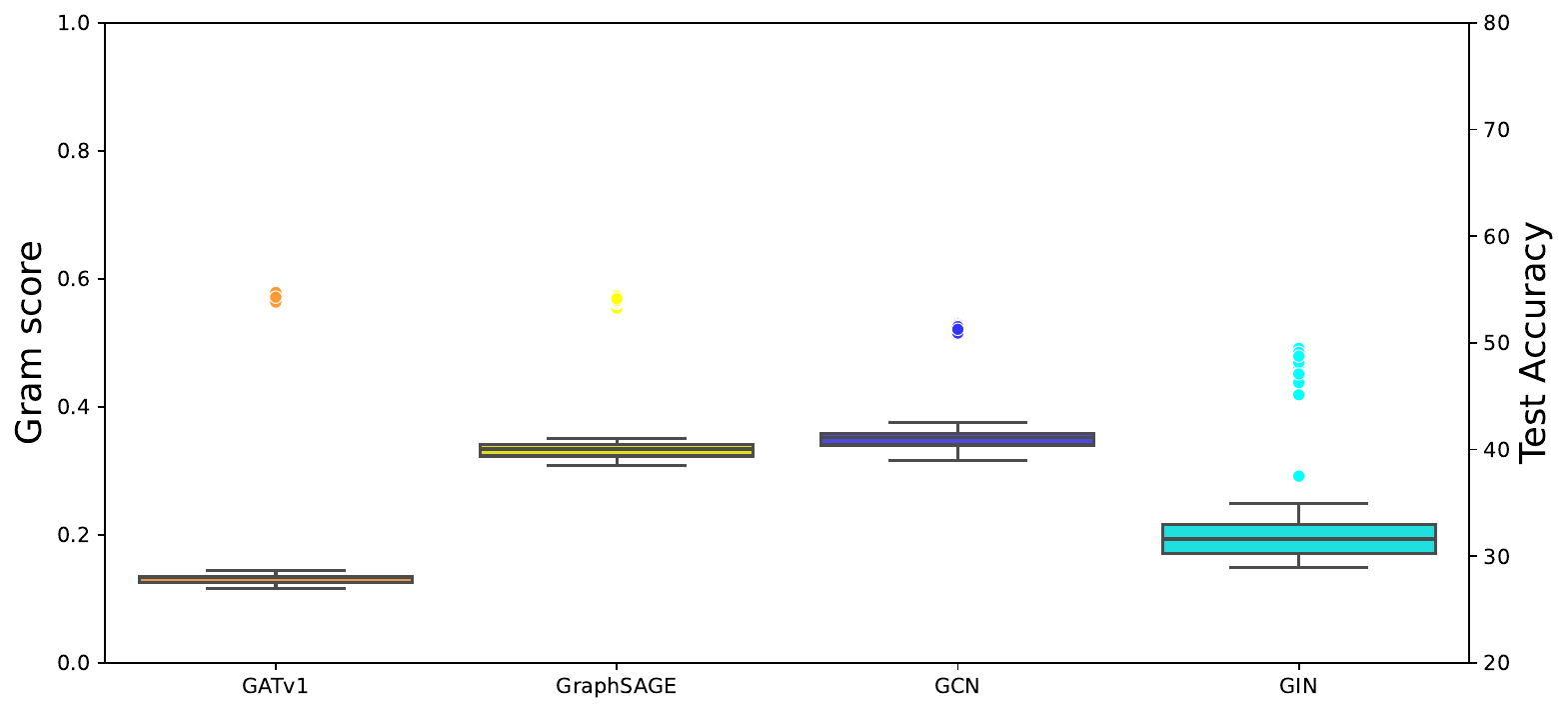}\label{fig:subfig4}}
    \subfigure[roman-empire]{\includegraphics[width=0.48\linewidth]{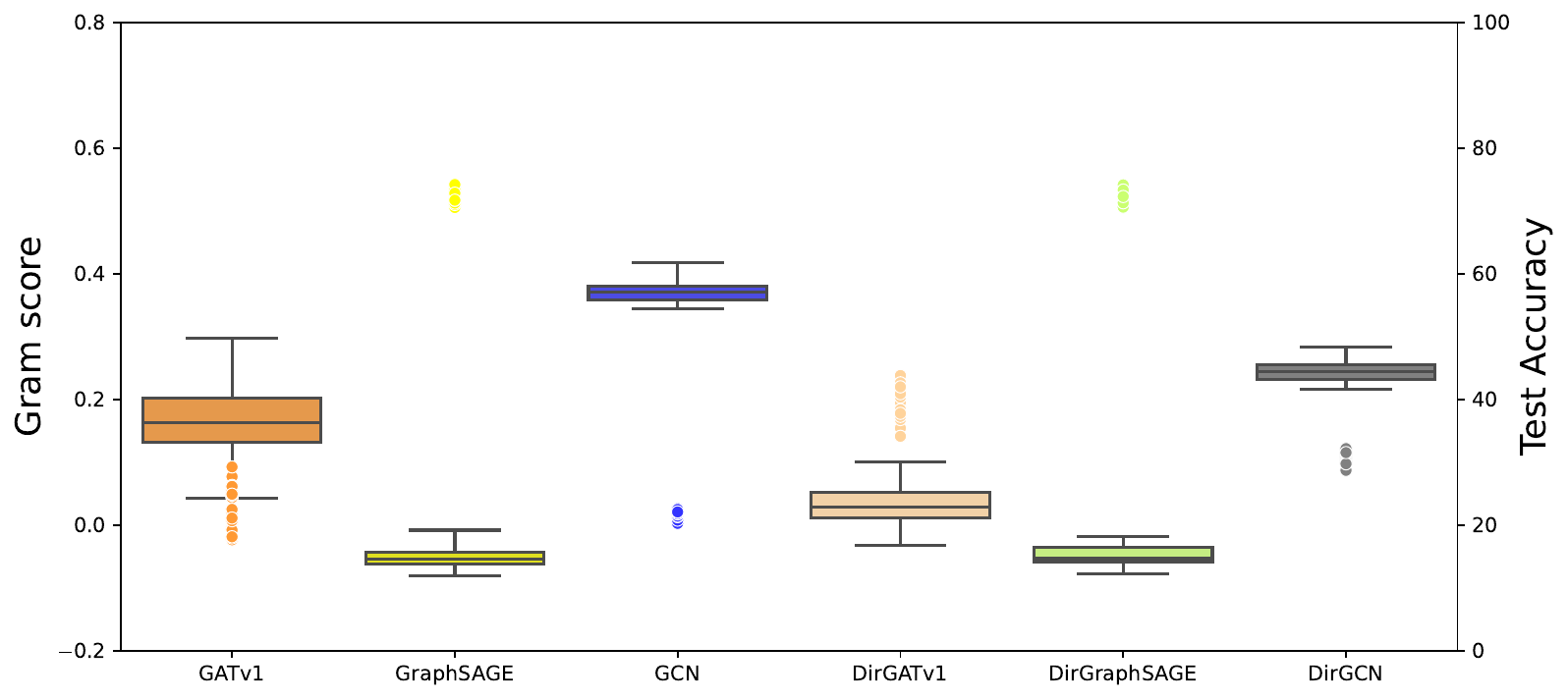}\label{fig:subfig5}}
    \subfigure[roman-empire 20-Jaccard]{\includegraphics[width=0.48\linewidth]{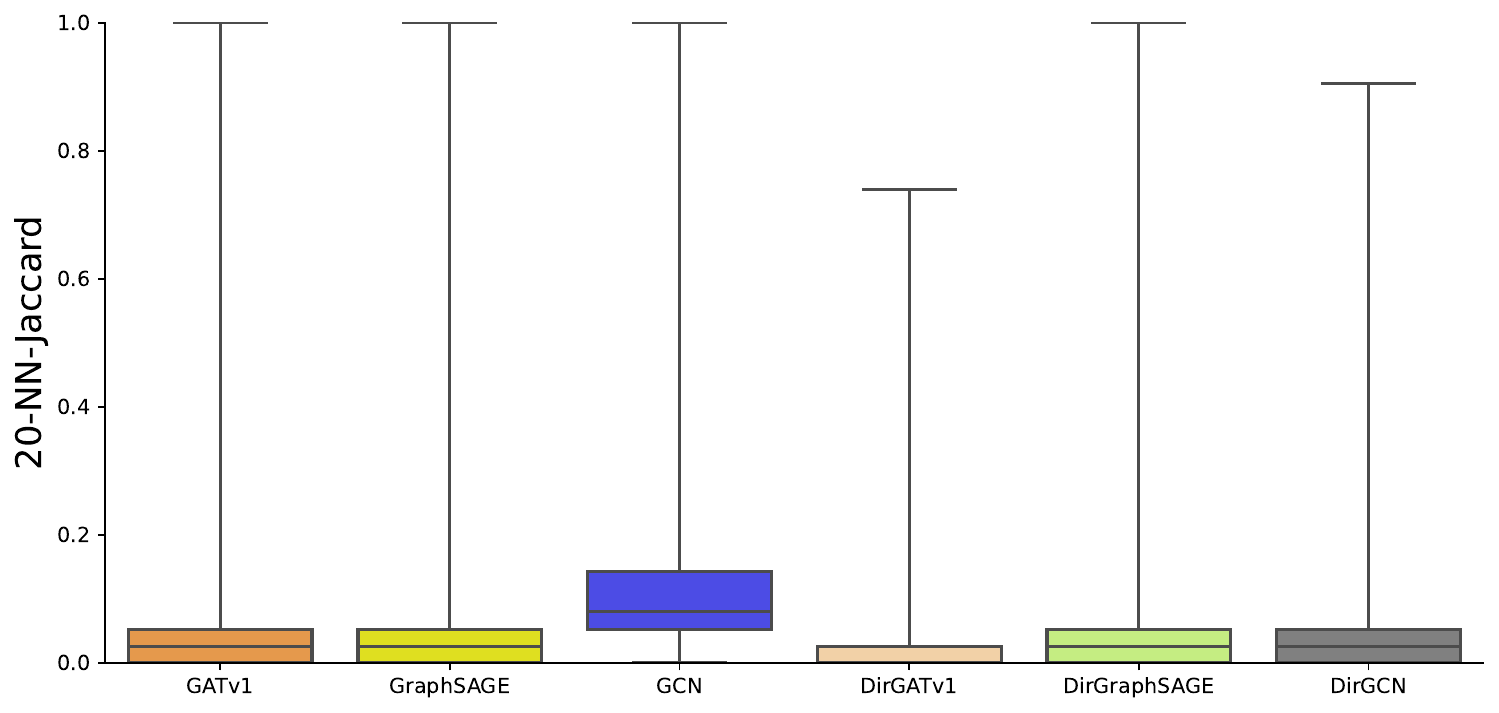}\label{fig:subfig6}}
    \caption{Stability and test accuracy plots. The box plots represent the distributions of summary Gram scores. The scatter plots represent test accuracies. The spread of the box plots and scatter plots represent GGI and variance of accuracies.}
\end{figure}

Models establish consistent stability behaviour for datasets of varying sizes~\ref{fig:subfig3}~\ref{fig:subfig4}, but GATv1 shows contrasting stability behaviour for homophilic and heterophilic datasets~\ref{fig:subfig5}. The impact of other dataset characteristics such as degree distribution remains as an open problem.

There is some evidence of a positive correlation between geometric and downstream stability. There also seems to be a weaker correlation between geometric stability and accuracy itself. Intuitively, this suggests a `good and consistent' model for a dataset learns a fixed (up to equivariance) shape of point clouds. A more detailed ablation study of the impact of architectural (e.g number of layers) and hyperparameter choices (e.g embedding dimension) is required to draw a stronger conclusion.

\setlength{\abovecaptionskip}{2pt}
\setlength{\belowcaptionskip}{2pt}

\begin{wraptable}{r}{0.5\textwidth}\label{fig:tableggi}
\begin{tabular}{|c|c|c|c|}
   \hline
      & Cora & ogbn-arxiv & roman-empire \\
    \hline
    GATv1 & 1.7 &  0.7 &  6.7 (3.0) \\
    \hline
    GraphSAGE &  3.0 & 1.2 & 1.7 (1.6) \\
    \hline
    GCN & 4.0 & 1.6 & 1.8 (1.6) \\
    \hline
    GIN & 6.7 & 3.1 &  \\
    \hline
\end{tabular}
  \caption{GGI for given models and datasets.}
\end{wraptable}

The difficulties of applying previous indices also become apparent. For ogbn-arxiv, it takes over a day compute 20NN-Jaccard, and prohibitively long for all other indices. It however only takes \textbf{less than 20 minutes} to calculate GGI with a single GPU. The lack of invariance properties makes it hard to obtain conclusive results. 20NN-Jaccard is completely uninformative for roman-empire dataset ~\ref{fig:subfig6}. A confounding cause could be that measuring neighbouring embeddings is less effective for a heterophilic dataset.
 Moreoever, 20NN-Jaccard and second order cossim give contradictory results for cora (Appendix~\ref{app:ncresults}).

\textbf{Link prediction}
To the best of our knowledge there's no previous literature studying the geometric stability of embeddings produced by link prediction models. We evaluate batched versions of GraphSAGE and GCN (with GraphSAINT sampling\cite{GraphSAINT}) on ogbl-citation2\cite{ogb}. Standard training setup is detailed in Appendix~\ref{app:experimentsetup}.

We observe GGI for batched GraphSAGE and GCN to be \textit{2.5\%, 2.6\%} respectively with box plot in Appendix~\ref{app:experimentsetup} Figure~\ref{fig:lpplots}. Interestingly, both GGI index and box-plot statistics align with the observations we see during full-batch training of respective models, on different datasets and even tasks.

\bibliographystyle{unsrtnat}
\bibliography{reference}

\section*{Acknowledgements}
We thank all reviewers for their helpful review, including the suggestions of adding a heterophilic dataset and coplotting downstream accuracies.

\appendix
\section{Previous geometric stability indices}\label{app:indices}
\textbf{Aligned cosine.}
 Given two set of embeddings $Z^l, Z^m$, solve the orthogonal Procrustes problem to obtain $Q^{l,m} \in \mathbb{R}^{d \times d}$. For a given node $i$, $s_i^{l,m}$=$cossim(s_{i}^l Q^{l,m}, s_{i}^m)$. The aligned cosine stability index is the average over all nodes and configuration pairs.

\textbf{kNN-Jaccard.}
 For a given node $i$ and configuration $l$, find it's k-nearest neighbours in $l$ and $m$. The kNN-Jaccard overlap $s(z_i^l, z_i^m)$ = $\frac{N_k^l(v_i) \cap N_k^m(v_m)}{N_k^l(v_i) \cup N_k^m(v_m)}$. The kNN-Jaccard stability index is the average over all nodes and configuration pairs.

\textbf{Hausdorff distance.}
Given two embeddings $Z^l$, $Z^m$, the Hausdorff distance is $d_H(Z^l, Z^m) = max \{{\sup\limits_{z_i \in Z^l} d(z_i,Z^m), \sup\limits_{z_i \in Z^m} d(z_i, Z^l)}\}$. The Hausdorff distance stability index is the average over all configuration pairs.

\section{Time and space complexities}\label{app:complexities}
\textbf{Aligned cosine time complexity.}
To solve orthogonal Procrustes of $Z^l$ and $Z^m$ where $Z$ has dim $|V|*d$, it's equivalent to solve SVD of $Z^l * Z^{mT}$ which takes $O(|V|^3)$.
Then for all pairs of runs, it takes $O(N^2|V|^3)$.
Then finding $s^{acs}(z_i^, z_m^l)$ = $cos(z_i^lQ^{l,m}, z_i^m)$ takes $O(d^2)$ for multiplication (and $O(d)$ for cos). For all seed pairs for all nodes, it takes $O(|V|N^2d^2)$, adding up to $O(N^2|V|^3) + O(|V|N^2d^2)$.

\textbf{Aligned cosine space complexity.}
Remember all $Q^{l,m}$ takes $O(N^2d^2)$.

\textbf{kNN-Jaccard time complexity.}
For kNN for a vector $v_i$, $O(|V|d)$. All kNN neighbours for all nodes, across all runs is $O(|V|^2Nd)$.
For a node $v_i$, to find similarity between it's two embeddings for two runs $l$, $m$, $s(z_i^l, z_i^m)$ = $\frac{N_k^l(v_i) \cap N_k^m(v_m)}{N_k^l(v_i) \cup N_k^m(v_m)}$, takes times $O(k)$. Over all nodes and configuration pairs it takes $O(|V|^2Nd + |V|N^2k)$.  

\textbf{kNN-Jaccard space complexity.}
All neighbour lists for all nodes across runs, $O(|V|Nk)$ number of nodes. All embeddings take $O(Nd|V|)$ which adds up to $O(|V|Nk) + O(Nd|V|)$.

\textbf{Second-order cossim time complexity.}
kNN for all nodes across all runs, $O(|V|^2Nd)$. 
For each node $v_i$, find $N_k^l(v_i) \cup N_k^m(v_i)$ in $O(k)$. Then $s_j^l(v_i) = cossim(z_i^l, \phi^l(u_j))$ for each $u_j$ in unioned neighrbous takes $O(d)$ for one $u_j$ and $O(dk)$ altogether. $s^m(v_i)$ in $O(dk)$ time as well. 
Then to compute $s_k^{cos}(z_i^l, z_i^m) = cossim(s^l, s^m)$ takes $O(k)$, the length of two vectors. So altogether $O(dk+k)=O(dk)$.
Repeat this for all pairs across configurations and all nodes takes $O(dkN^2|V|)$ time.
Therefore $O(|V|^2Nd + dkN^2|V|)$ overall.

\textbf{Second-order cossim space complexity.}
All neighbour lists for all nodes across runs, $O(|V|Nk)$ number of nodes. All embeddings of all runs need to be stored, $O(Nd|V|)$.
Each node $v_i$ also need to store it's second order similarity vector $s_l(v_i)$ for each run $l$, so $O(kN|V|)$ float numbers.
Added up it's $O(|V|Nk) + O(Nd|V|) + O(kN|V|)$.

\textbf{Hausdorff distance time complexity.}
Finding (max) of supremum over all configuration pairs takes $O(|V|^2N^2d)$.

\textbf{Wasserstein distance time complexity.}
All bijections between sets of size $|V|$ can be computed in $|V|^2$. Then summing the norm of differences takes $O(|V|d)$ time. Over all $|V|^2$ bijections it takes $O(|V|^3d)$ time. 

\section{Orthogonal Procrustes Problem}\label{app:procrustes}
Given two matrix $X$, $Y$, the orthogonal Procrustes problem finds an orthogonal matrix $\Omega$ that best transforms $X$ into $Y$. It solves $e = \argmin_{\Omega} \lVert XT - Y \rVert_F$. The problem reduces to computing SVD of $YX^T$ which takes time $\mathcal{O}(n^3)$.

\section{Hausdorff Distance}\label{app:hausdorff}
Given to subsets $X$, $Y$ of a metric space $(M, d)$, the Hausdorff distance $d_H(X,Y) = max \{{\sup\limits_{X \in X} d(x,Y), \sup\limits_{y \in Y} d(y, X)}\}$. When $X$, $Y$ are compact, $d$ lifts to a proper metric $d_H$ over the subsets. Intuitively, this is because `Hausdorffness' provides the level of separation and compactness guarantees the density of open sets together allows the notion of distances.

The most natural application of Hausdorff distance in computer science is for matching (two) subimages\cite{hausdorff}. This however does not apply for graph embeddings. Each point in the embedding spaces corresponds to a specific node and the distance between $z^l_{v_i}$ and an arbitrary farthest $z^m_{v_j}$ does not immediately measure any geometric embedding stability. Moreover the choice of Hausdorffness (in the spectrum of separation axioms, $T_1$) in our context is somewhat arbitrary. We do not require all the nice properties a Hausdorff topological space has about the space's limits and metrizability. Some definition of $T_0$-distance or $T_{3\frac{1}{2}}$ (Tychonoff)-distance could have been defined as well.

\section{Centered Kernel Alignemnt}\label{app:cka}
CKA\cite{CKA} addresses the problem of measuring similarity of the internal representations of two different neural networks (across datasets) that extends previous methods such as canonical correlation analysis\cite{SVCCA}.

We draw attention the below extremely simple equation which they use, for matrix $X \in \mathbb{R}^{N \times d_1}, Y \in \mathbb{R}^{N \times d_2}$

\begin{align*}
    \langle \text{vec}(XX^T) , \text{vec}(YY^T) \rangle &= \text{tr}(XX^TYY^T) = \lVert Y^TX \rVert^2_F
\end{align*}

While algebraically simple, this crucially removes computation between $Y$ and $X$, at the cost of computing within $X$ and $Y$ independently. This methodology is also one of our motivations for using the Gram matrix and separating GGI from previous indices by following the different path in the commutative diagram.

\section{Proof of invariance properties of GGI}\label{app:proof}
\begin{proof*}
\textit{Node permutation:} Given a permutation $\sigma: V \rightarrow V$ of nodes which induces a bijection on edges $\sigma: E \rightarrow E$. $s^l = \frac{1}{2|E|}\sum_{i,j \in |V| \times |V|}S^l[i,j] =  \frac{1}{2|E|}\sum_{i,j \in E} (ZZ^T)[i,j] = \frac{1}{2|E|}\sum_{\sigma(i),\sigma(j) \in E} (ZZ^T)[\sigma(i),\sigma(j)]$.

\textit{Orthogonal transformation and translation:} $s^l  = \frac{1}{2|E|}\sum_{i,j \in E} (ZZ^T)[i,j] = \sum_{i,j \in E} \langle z_i, z_j\rangle = \sum_{i,j \in E} \langle Tz_i , Tz_j \rangle$ for any isometry $T \in \mathbb{R}^{d \times d}$.

Once we obtain $s^l$ scores for each embedding, any final aggregation into a single index can only be a variant to the order of evaluation. Taking standard deviation is not dependent on the ordering.
\end{proof*}

\section{20NN-Jaccard and second-order cossim node classification results}\label{app:ncresults}
\begin{figure}[htbp]
    \centering
    \subfigure[20-NN-Jaccard on Cora.]{\includegraphics[width=0.45\linewidth]{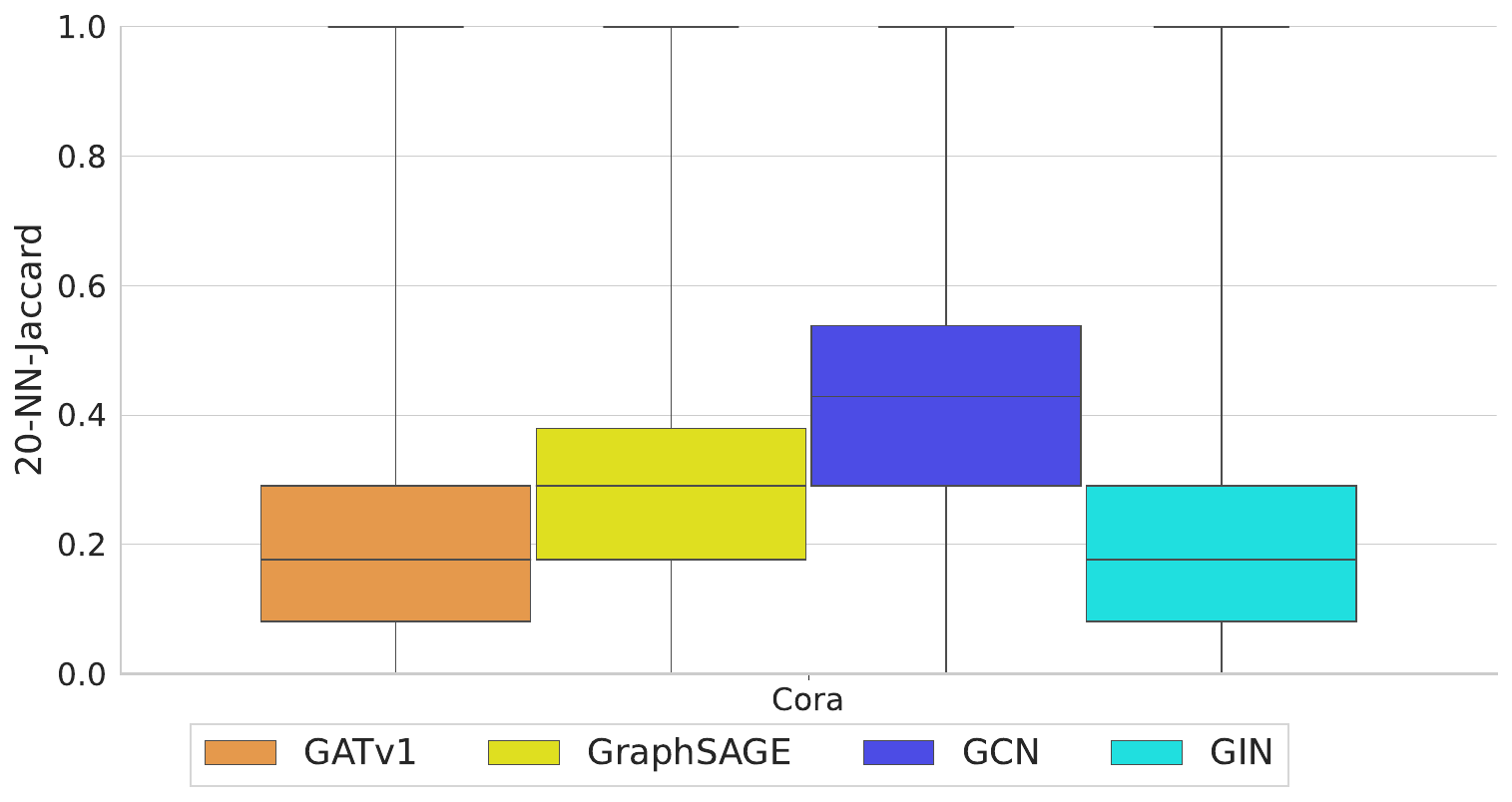}\label{fig:subfig1}}
    \hfill
    \subfigure[Second-order cossim on Cora.]{\includegraphics[width=0.45\linewidth]{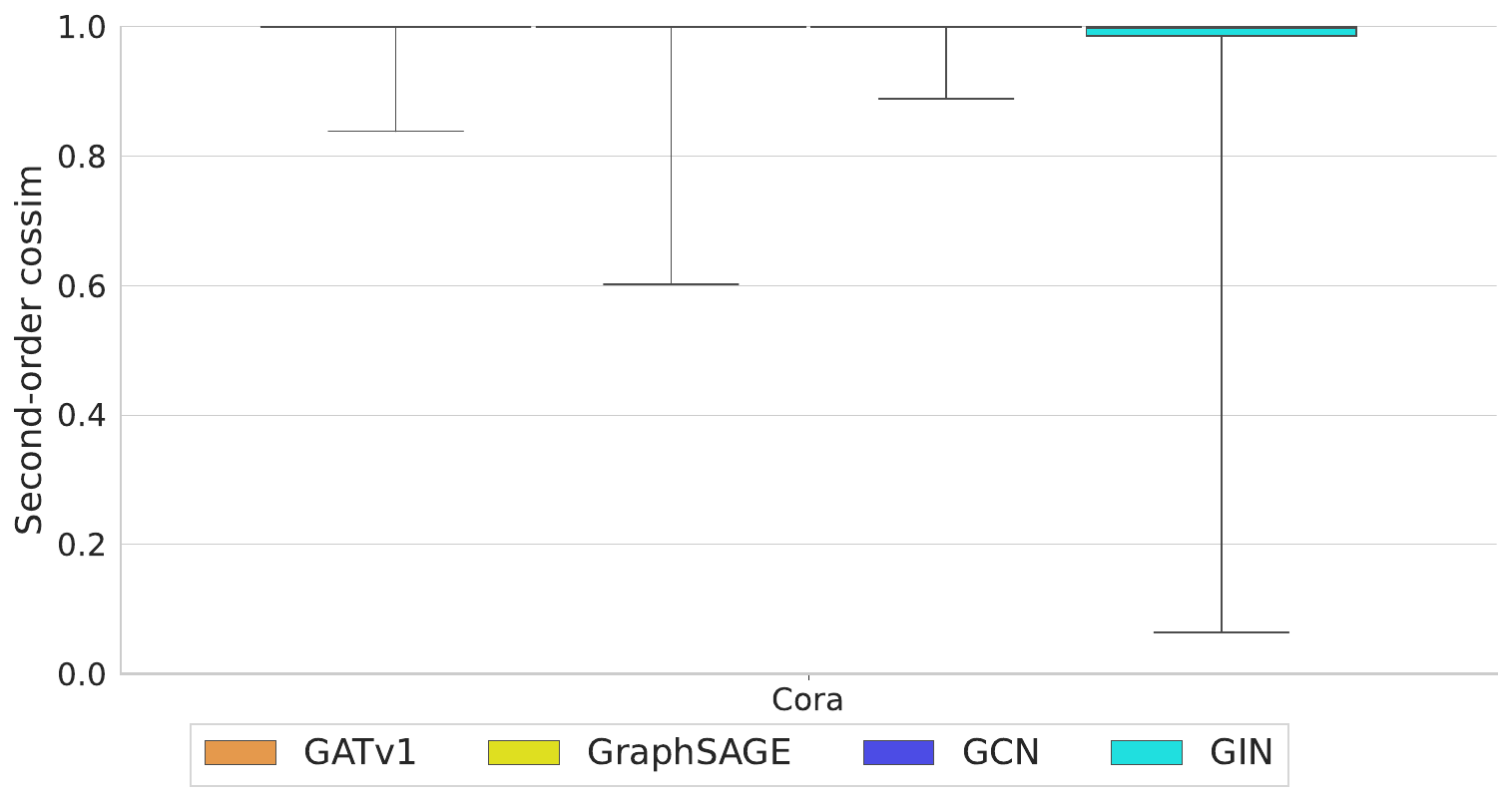}\label{fig:subfig2}}
    \label{fig:cora-old}
    \caption{Box plots of previous stability indicies on Cora. The second-order cossim is uninformative.}
\end{figure}

\ref{fig:subfig1} shows embeddings of GATv1 and GIN are less stable as compared to GraphSAGE and GCN. \ref{fig:subfig2} Second order cossim is however uninformative, as illustrated by \ref{fig:subfig2}, where all scores are close to1.0, with differences only seen in the min values. This further shows the difficulty of drawing consistent conclusions based on these methods.

\section{Experimental setup}\label{app:experimentsetup}
Code available at: \url{https://github.com/brs96/geometric-instability-gnn-large-graphs}

\textbf{Node classification.}
We use a machine with 60GB of vRAM and 8 vCPU, with.one NVIDIA T4 GPU. 

For training: Epochs = 100 and embedding dimension = 64 on Cora. Epochs = 500 and embedding dimension = 128 on ogbn-arxiv. Learning rate = 0.01. Number of layers = 3  (hence embeddings produced by the second layer). Loss = cross entropy. Model training implemented with PyG\cite{pyg}. PyG loaders are used for dataset splits for cora (\textit{Planetoid}, split from \cite{yang2016revisiting}) and ogbn-arxiv (by year). For roman-empire the split is from \textit{HeterophilousGraphDataset} where one of the 10 provided random splits are (randomly) picked for each training setup.

\textbf{Link prediction.}
We follow the training setup from batched GraphSAGE and GCN(GraphSAINT). We use a machine with 200GB of vRAM for ogbl-citaiton2. Equal number of positive and negative samples are used for GraphSAGE batches.Binary cross entropy loss of equal positive and negative weights are used. Embedding dimension = 128. Epochs = 200. Learning rate = 0.001. Batch size = 512. We evaluate \textbf{10} random seeds due to time and resources required for training on ogbl-citation2. For GraphSAINT sampling, walk length = 3 and number of steps = 100. Dataset split given by PyG ogbl-citation2 loader is used.

\begin{figure}[htbp]
    \centering
     \includegraphics[width=0.45\linewidth]{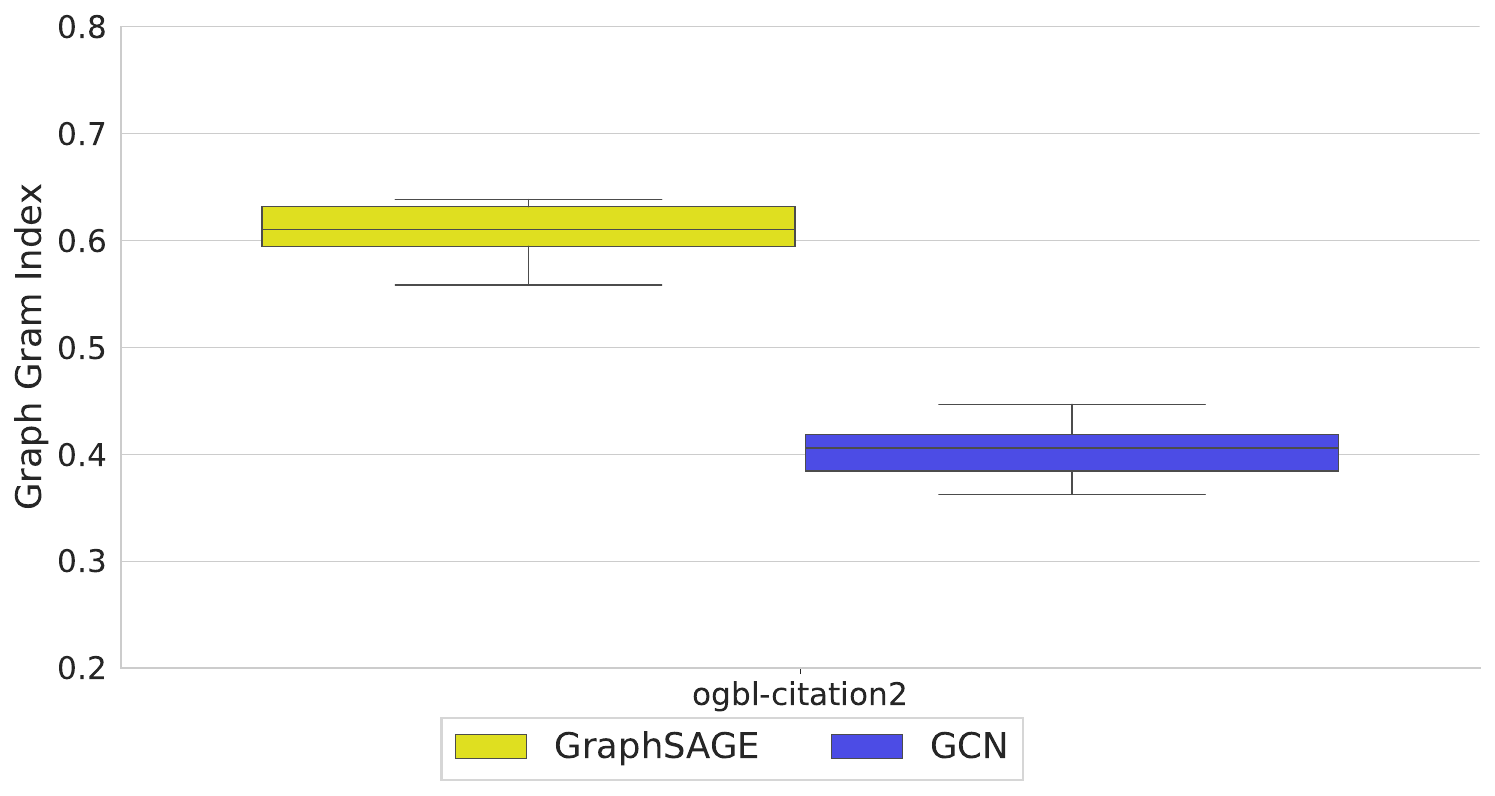}
    \hfill
    \caption{Box plots of GGI on ogbl-citation2.}
    \label{fig:lpplots}
\end{figure}

\section{Future work}\label{app:futurework}
\textbf{Graph Gram Index.}
The framework of GGI is flexible enough to allow various extension, for example by using non-Gram matrix based summary structure, using other the $S^l \rightarrow s^l$ pooling step such as the Frobenius norm (notably, this step is the same as many graph pooling steps), and replacing std with other summary statistics. 

\textbf{Invariance properties.}
What kind of invariance properties earlier stability indices satisfy is unstudied. It is also unknown whether the proven set of invariances for GGI (node permutation, orthogonal transformation and translation) is the complete set.

Meaningfully measuring stability under specific equivariance\cite{satorras2022en} or the sensitivity of any index to a particular transformation (for example, subspace translation of a one class\cite{davari2022reliability}) remains unexplored open questions.
It would be an interesting future direction to understand what type of instability is captured by GGI, and by any other index that can be created following our framework.

\end{document}